\documentclass[sigplan,screen,nonacm]{acmart}
\renewcommand\footnotetextcopyrightpermission[1]{} 
\AtBeginDocument{%
  \providecommand\BibTeX{{%
    \normalfont B\kern-0.5em{\scshape i\kern-0.25em b}\kern-0.8em\TeX}}}

\begin{document}

\title{Multimodal Learning for Cardiovascular Risk Prediction using EHR Data}

\author{Ayoub Bagheri}
\additionalaffiliation{%
  \institution{Department of Cardiology, University Medical Center Utrecht}
  \city{Utrecht}
\country{The Netherlands}
}
\email{a.bagheri@uu.nl}
\affiliation{%
  \institution{Methodology and Statistics \\ Utrecht University}
  \city{Utrecht}
\country{The Netherlands}
}

\author{T. Katrien J. Groenhof}
\email{t.k.j.groenhof@umcutrecht.nl}
\affiliation{%
  \institution{Julius Center for Health Sciences and Primary Care \\ University Medical Center Utrecht}
\city{Utrecht}
\country{The Netherlands}
}

\author{Wouter B. Veldhuis}
\email{w.veldhuis@umcutrecht.nl}
\affiliation{%
\institution{Department of Radiology \\ University Medical Center Utrecht}
  \city{Utrecht}
\country{The Netherlands}
}

\author{Pim A. de Jong}
\email{pjong8@umcutrecht.nl}
\affiliation{%
\institution{Department of Radiology \\ University Medical Center Utrecht}
  \city{Utrecht}
\country{The Netherlands}
}

\author{Folkert W. Asselbergs}
\email{f.w.asselbergs@umcutrecht.nl}

\additionalaffiliation{%
  \institution{Institute of Cardiovascular Science, University College London}
  \city{London}
\country{United Kingdom}
}

\additionalaffiliation{%
  \institution{Health Data Research UK and Institute of Health Informatics, University College London}
  \city{London}
\country{United Kingdom}
}

\affiliation{%
\institution{Department of Cardiology \\ University Medical Center Utrecht}
  \city{Utrecht}
\country{The Netherlands}
}

\author{Daniel L. Oberski}
\email{d.l.oberski@uu.nl}

\additionalaffiliation{%
  \institution{Julius Center for Health Sciences and Primary Care, University Medical Center Utrecht}
  \city{Utrecht}
\country{The Netherlands}
}

\affiliation{\institution{Methodology and Statistics \\ Utrecht University}
\city{Utrecht}
\country{The Netherlands}}

\renewcommand{\shortauthors}{Bagheri, et al.}

\begin{abstract}
  Electronic health records (EHRs) contain structured and unstructured data of significant clinical and research value. Various machine learning approaches have been developed to employ information in EHRs for risk prediction. The majority of these attempts, however, focus on structured EHR fields and lose the vast amount of information in the unstructured texts. To exploit the potential information captured in EHRs, in this study we propose a multimodal recurrent neural network model for cardiovascular risk prediction that integrates both medical texts and structured clinical information. The proposed multimodal bidirectional long short-term memory (BiLSTM) model concatenates word embeddings to classical clinical predictors before applying them to a final fully connected neural network. In the experiments, we compare performance of different deep neural network (DNN) architectures including convolutional neural network and long short-term memory in scenarios of using clinical variables and chest X-ray radiology reports. Evaluated on a data set of real world patients with manifest vascular disease or at high--risk for cardiovascular disease, the proposed BiLSTM model demonstrates state-of-the-art performance and outperforms other DNN baseline architectures.
\end{abstract}




\keywords{clinical text mining, text classification, cardiovascular risk prediction, clinical multimodal learning}

\maketitle

\section{Introduction}
Electronic health records (EHRs) data have become increasingly available to researchers as more hospitals, clinics and practices have adopted data digitization. EHRs store data in different modalities, such as structured tabular data (e.g. demographic values, laboratory results, medications) and unstructured texts (e.g. referral letters, clinical notes, discharge summaries, radiology reports). This digitization creates an opportunity to mine the health records to increase the quality of care and clinical outcomes. Yet clinicians have limited time to process all the available data and detect patterns across similar medical records. Deep learning and machine learning, on the other hand, are suitable for discovering useful patterns from vast amount of data.

Unstructured texts contained within the EHRs are recognized as a rich but not easily accessible and usable source of medical information \cite{sheikhalishahi2019, drozdov2020, zech2018, wang2012}. Recent studies have attempted to derive information from unstructured medical texts to classify disease codes \cite{du2019, bagheri2020b}, detect patient's disease history \cite{bagheri2020, wu2020}, and predict hospital readmission or clinical outcomes \cite{alex2019, huang2019, jagannatha2016}. X-ray radiology reports are example of such unstructured data describing radiologist's observations on patient's medical conditions associated to medical images. The majority of previous decision support systems for radiology reports are developed using rule-based approaches applied on unstructured and semi-structured texts \cite{gong2008, chen2018, taira2001, wang2018}. However, these methods are often impractical because they do not generalize to new data and often are not applicable for big data analysis \cite{pons2016}. 

Recent studies have shown promising results using free-text radiology reports and deep learning models to predict clinical outcomes \cite{monshi2020, chen2018, laserson2018, smit2020, wood2020, shin2017}. Convolutional neural networks (CNNs) and recurrent neural networks (RNNs) are two common deep learning techniques that have been effective in text mining and natural language processing (NLP), also in EHR applications \cite{jagannatha2016, monshi2020, du2019, banerjee2019}. Deep learning-based modelling of radiology reports has been proposed to supersede the simple grammatical patterns and hand-crafted regular expressions of the traditional clinical rules-based software, such as PEFinder \cite{chapman2011}, medical language extraction and encoding system (MedLEE) \cite{hripcsak2002, friedman1994} and CTakes \cite{savova2010}. While these neural networks models gained tremendous momentum in knowledge discovery from EHR texts, there are very seldom studies that used of both free-texts and the structured information in EHRs for clinical prediction and classification \cite{scheurwegs2016, liu2018, xu2018, jin2018}.

In this paper we leverage structured features in EHR data, i.e. lab results, to combine with free-text radiology reports to uncover patterns to improve cardiovascular risk prediction. The primary contribution of this study is twofold:

\begin{itemize}
    \item We present a multimodal bi-directional long short-term memory (BiLSTM) neural network that integrates the neural text representation with laboratory results and feeds them into a fully connected neural network.
    \item We investigate the effectiveness of the proposed architecture to predict cardiovascular risk for real world patients with manifest vascular disease or at high-risk for cardiovascular disease \cite{simons1999} visiting the University Medical Center (UMC) Utrecht.
\end{itemize}

The rest of paper is organized as follows, section 2 provides related work in the literature, section 3 details the proposed multimodal architecture for mining EHR data. Experiments and results are given in section 4. Section 5 concludes this research.

\section{Related Work}
Many text mining pipelines and NLP systems have been developed to extract structured information from free-text radiology reports. These range from simple rule-based, regular expression, and bag-of-word methods, to more sophisticated machine learning- and deep learning-based approaches \cite{zech2018, wood2020, taira2001, smit2020, shin2017, sevenster2012, pons2016, monshi2020, laserson2018, gong2008, friedman1994, drozdov2020, cornegruta2016, chen2018, bustos2019, banerjee2019, xu2016}.

Sevenster et al. \cite{sevenster2012} evaluated a system that extracts and correlates clinical findings and body locations from radiology reports using MedLEE, a rule-based NLP-based system. Khalifa and Meystre \cite{khalifa2015} built a NLP application based on the Apache UIMA (Unstructured Information Management Architecture) and reusing existing tools previously developed. Using this application they addressed identifying risk factors for heart disease based on the automated analysis of narrative clinical records of diabetic patients. Drozdov et al. \cite{drozdov2020} applied five supervised machine learning algorithms on chest radiology reports; K-nearest neighbour, logistic regression, Gaussian naïve Bayes classifier, random forest, and support vector machine. These methods were evaluated on a term frequency-inverse document frequency matrix of radiology reports that was then transformed to lowered dimensions using singular value decomposition. These rule-based and traditional machine learning approaches are in the need of domain expertise and hard core feature extraction \cite{pons2016, sheikhalishahi2019, wang2018}.

Deep learning techniques have already shown potential to automate the task of classifying X-ray reports in a way that could inform decisions regarding medical utilization \cite{drozdov2020, chen2018, shin2017}. Recently, a supervised learning approach using a RNN model with attention mechanism achieved high accuracy on expert-labeled chest X-ray radiology reports data set \cite{bustos2019}. Similarly, Cornegruta et al. \cite{cornegruta2016} proposed a BiLSTM neural network which was demonstrated to perform favourably in a corpus of radiology reports. Drozdov et al. \cite{drozdov2020}, besides non-neural classifiers, applied LSTM and BiLSTM networks on chest X-ray radiology reports that could produce state-of-the-art classification results. Wood et al. \cite{wood2020} developed ALARM, a transformer-based network report classifier on MRI data using BioBERT \cite{lee2020} models trained on radiology reports, and demonstrated improvement over simpler word embedding methods \cite{zech2018, mikolov2013, pennington2014}. Smit et al. \cite{smit2020} demonstrated superior performance of a deep learning method for radiology report labeling, in which a biomedically pretrained BERT model is first trained on annotations of a rule-based labeler, and then finetuned on a small set of expert annotations. Finally, Chen et al. \cite{chen2018} implemented CNNs to extract pulmonary embolism findings from thoracic computed tomography reports, outperforming state-of-the-art NLP systems.

However, current studies on text mining for radiology reports lack the flexibility to utilize the benefits of using multimodal data, i.e. both unstructured text and structured numerical values. One study using both laboratory features and text reports is that of Liu et al. \cite{liu2018}. Liu et al. concatenated lab and demographic features to the output features of medical notes obtained from a max-pooling layer of their neural network model. In their study, CNN- and RNN-based deep learning models have been developed for chronic disease prediction using medical notes, and  then a fully connected neural network has been applied with one hidden layer at the end of the pipeline. Another similar research is by Xu et. al \cite{xu2018}, in which separate machine learning models were trained with data from unstructured text, semi-structured text and structured tabular data to create a model that predicts diagnostic codes of international classification of diseases (ICD-10). 

In this paper, we present a multimodal learning architecture with a bidirectional deep learning-based model for free-text radiology reports. The bidirectional model can deal with contextual dependencies in text as it trains two LSTM networks on each side of the current word, running from left to right and from right to left, to encode and represent the text. The research presented in this paper is different from previous research as it: (1) combines structured laboratory results and unstructured text in a machine learning model to make an accurate decision of classifying cardiovascular events. (2) uses real world EHR data including X-ray reports in Dutch for cardiovascular risk prediction.

\section{Methodology}
In this section, we describe the details of our proposed methods, including the used cohort study, data ethics and privacy, text preprocessing, and the employed multimodal recurrent neural network.

\subsection{Cohort Study}
The patients included in this study originated from the Second Manifestations of ARTerial disease (SMART) study. The design of the SMART study is published elsewhere \cite{simons1999}. In short, the SMART study is an ongoing single-center prospective cohort study that was designed to establish the presence of additional arterial disease and risk factors for atherosclerosis in patients with manifest vascular disease or a vascular risk factor. Patients visiting the UMC Utrecht for evaluation of an atherosclerotic cardiovascular condition are eligible for inclusion in SMART. Inclusion criteria are presentation with an atherosclerotic cardiovascular condition and age > 18 years. Exclusion criteria are life expectancy < 3 months, unstable vascular disease and insufficient fluency in the Dutch language. A total of 5603 patients and their reports were included in this analysis. Characteristics are shown in Table 1. Variables that are predictors in the SMART risk score (age, sex, smoking, systolic blood pressure, diabetes, high-density-lipoprotein (HDL) cholesterol, total cholesterol, renal function according to the modification of diet in renal disease (MDRD) formula, history of cardiovascular disease stratified for stroke, peripheral artery disease, abdominal aortic aneurysm, and coronary heart disease, and years since diagnosis of first cardiovascular disease) were extracted for all patients. MACE (MAjor Cardiovascular Events) during followup is the outcome variable for prediction. Missingness of data was solved using the MICE package \cite{buuren2010} with one imputation for each missing value.

\begin{table}
\caption{\textbf{Patients' Characteristics in the SMART Study}}
\label{tab:my-table}
\begin{tabular}{lc}
\toprule
\textbf{Characteristic}                                     & \textbf{Total n = 5603}   \\
\midrule
\textbf{Age},   years (mean (sd))                           & 56.2 (12.5)      \\
\textbf{Female   sex}, n (\%)                               & 1926 (34.4)        \\
\textbf{Current   smoking}, n (\%)                          & 1549 (27.6)        \\
\textbf{History   of cardiovascular disease}                &                  \\
\hspace{4mm}CHD*, n (\%)                                    & 2166 (38.66)        \\
\hspace{4mm}Stroke, n (\%)                                 & 1076 (19.20)        \\
\hspace{4mm}PAD*, n (\%)                                    & 631 (11.26)       \\
\hspace{4mm}AAA*, n (\%)                                    & 306 (5.46)       \\
\hspace{4mm}Years since first diagnosis \\ 
\hspace{4mm}of CVD*,   median (IQR)                             & 0 (0-4)          \\
\textbf{Risk factors for cardiovascular disease}            &                  \\
\hspace{4mm}Diabetes Mellitus, n (\%)                      & 1047 (18.69)        \\
\hspace{4mm}Hypertension, n (\%)                           & 2353 (42.00)        \\
\hspace{4mm}Dyslipidemia, n (\%)                           & 432 (7.71)                  \\
\textbf{BMI*},   kg/m2 (mean (sd))                           & 26.8 (4.3)       \\
\textbf{SBP*},   mmHg (mean (sd))                            & 140 (21)         \\
\textbf{DBP*},   mmHg (mean (sd))                            & 83 (13)          \\
\textbf{Laboratory}                                         &                  \\
\hspace{4mm}Total cholesterol, mmol/L (mean (sd))              & 5.14 (1.38)      \\
\hspace{4mm}LDL*-cholesterol, mmol/L (mean (sd))                & 3.1 (1.16)       \\
\hspace{4mm}HDL-cholesterol, mmol/L (mean (sd)                 & 1.27 (0.38)      \\
\hspace{4mm}Triglycerides, mmol/L (median (IQR))               & 1.7 (1.2-2.5)    \\
\hspace{4mm}MDRD, ml/min/1.73m2 (median (IQR))                 & 80 (68-91)       \\
\hspace{4mm}HbA1c*, mmol/mol (median (IQR))                     & 5.7 (5.4-6.1)    \\
\hspace{4mm}Glucose, mmol/L (median (IQR))                     & 5.7 (2.6-6.4)    \\
\hspace{4mm}Hemoglobin, mmol/L (mean (sd))                     & 6.0 (2.04)       \\
\hspace{4mm}Creatinine, $\mu$mol/L (median (IQR))             & 84 (73-97)       \\
\hspace{4mm}CRP*, mg/L (median (IQR))                          & 1.95 (0.90-4.20) \\
\hspace{4mm}TSH*, mU/l (mean (sd))                            & 0.9 (0.09)       \\
\textbf{MACE during followup}, n (\%)                       & 1385 (24.72)       \\
\bottomrule
\end{tabular}
    \small
     {\raggedright * CVD: Cardiovascular disease, CHD: Coronary heart disease, PAD: Peripheral arterial disease, AAA: Abdominal aortic aneurysm, BMI: Body mass index, SBP: Systolic blood pressure, DBP: Diastolic blood pressure, LDL: Low-density lipoprotein, HbA1c: Hemoglobine A1c, CRP: C-reactive protein, TSH: Thyroid stimulating hormone  \par}
\end{table}

\subsection{Ethics and Privacy}
Informed consent was obtained through established procedures. The SMART study has been approved by the Medical Ethical Committee of the UMC Utrecht. All data are handled according to data protection and privacy regulations.

\begin{figure*}[h]
  \centering
  \includegraphics[width=\textwidth]{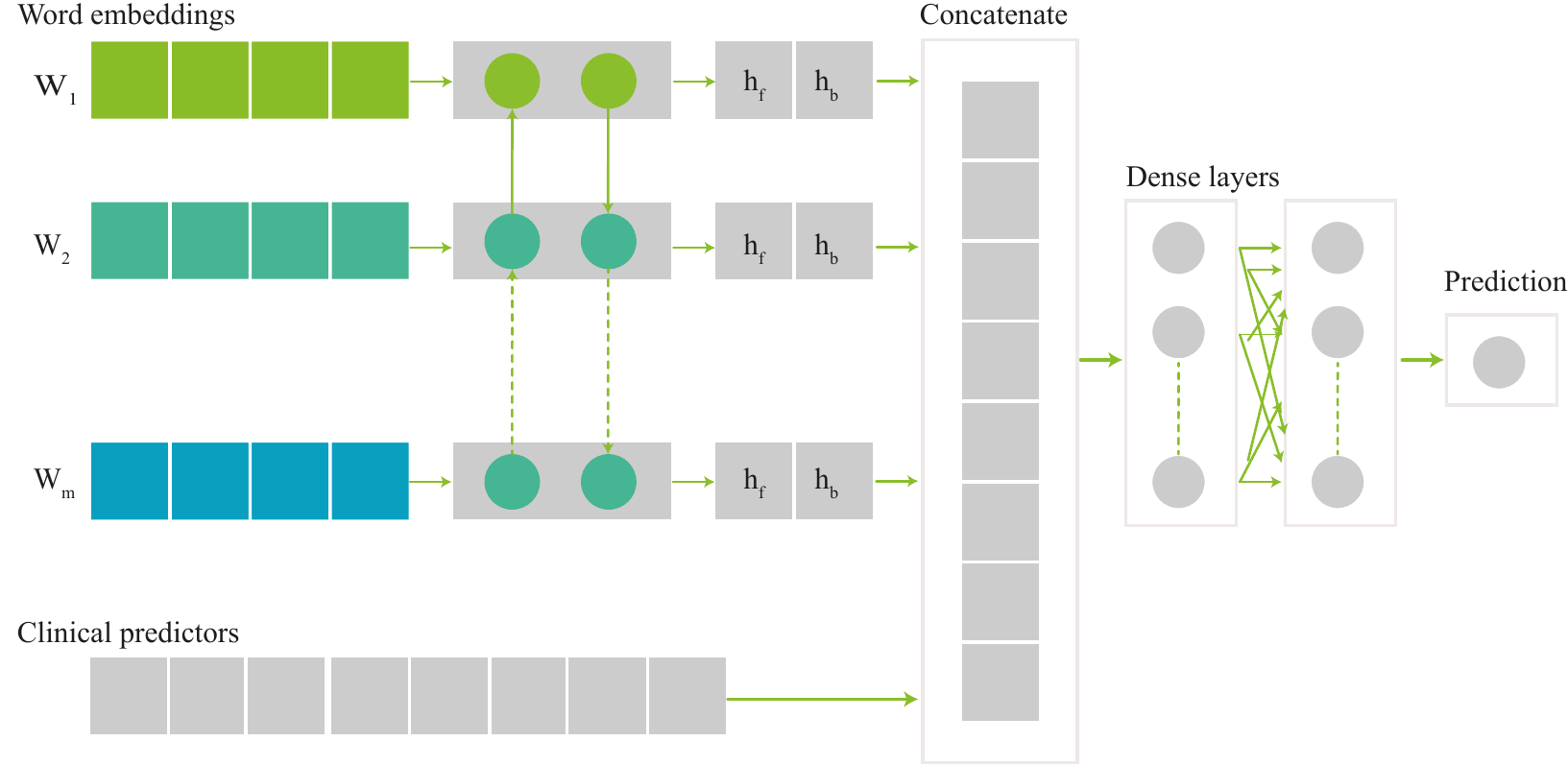}
  \caption{\textbf{Multimodal Recurrent Neural Network}}
\end{figure*}

\subsection{Text Preprocessing}
Radiology reports in the SMART study, for the process of text mining, may contain redundant characters and words such as punctuation marks and stop words. Therefore, we perform the following preprocessing steps to improve the quality of text data for the ongoing study: (1) All characters are transformed into lowercase. (2) We remove numbers and some meaningless punctuation marks such as semicolon and colon. (3) Stop words are then removed. (4) We then apply the Porter's stemming algorithm \cite{porter2001, kraaij1994} on texts.

\subsection{Multimodal Recurrent Neural Network}
Our proposed model consists of an embedding layer, a BiLSTM layer, dropout, a concatenation layer and dense layers. Figure 1 shows the architecture of the multimodal BiLSTM model that integrates word embeddings and clinical predictors. 

\subsubsection{Embedding Layer}
To extract the semantic information of radiology reports, each text is firstly represented as a sequence of word embeddings. Word embedding is an improvement over bag-of-word models where large sparse vectors were used to represent each word. On the contrary, in an embedding, words are represented by dense vectors where a vector represents the projection of the word into a continuous vector space \cite{mikolov2013, mikolov2013b}. Denote $s$ as an X-ray report with $m$ words and each word is mapping to a vector, then we have:

\begin{equation}
    s = [\vec{e}_1, \vec{e}_2, ..., \vec{e}_m]
\end{equation}

where vector $\vec{e}_i$ represents the vector of \textit{i}-th word
with a dimension of $d$. The vectors of word embeddings are concatenated together to maintain the order of words in a patient report.

\subsubsection{Bidirectional-LSTM Layer}
After the embedding layer, the sequence of word vectors is fed into a bidirectional LSTM layer to achieve another representation of radiology reports. Interest in incorporating a BiLSTM layer into the architecture of our model arises from their ability to learn long-term dependencies and contextual features from both past and future states \cite{schuster1997}. The BiLSTM layer calculates two parallel LSTM layers, a forward hidden layer and a backward hidden layer, to generate an output sequence $y$ as illustrated:

\begin{equation}
  h_{f_t} = \sigma(W_{xh_{f}}x_t + W_{h_{f}h_{f}}h_{f_{t-1}} + b_{h_{f}})
\end{equation}

\begin{equation}
  h_{b_t} = \sigma(W_{xh_{b}}x_t + W_{h_{b}h_{b}}h_{b_{t-1}} + b_{h_{b}})
\end{equation}

\begin{equation}
  y_t = W_{h_{f}y}h_{f_t} + W_{h_{b}y}h_{b_t} + b_{y}
\end{equation}

Here $\sigma$ is the sigmoid activation function, \begin{math} x_t \end{math} is a $d$-dimensional input vector at time step $t$, $W$ are the weight matrices, $b$ are bias vectors, and $h_f$, $h_b$ are the output of the LSTM forward and backward layers, respectively.

\subsubsection{Dropout}
Large neural networks trained on relatively small data sets can overfit the training data. Dropout provides a powerful method of regularizing that prevents the potential overfitting issue by randomly setting input elements to zero with a given probability of dropout rate \cite{srivastava2014, goodfellow2016}. In our multimodal RNN model, dropout and recurrent dropout are used with the BiLSTM layer.

\subsubsection{Concatenation Layer}
The concatenation layer, in the multimodal RNN model, takes the outputs of the BiLSTM layer and the clinical predictors from the SMART study as inputs and concatenates them along a specified dimension. In this layer there are no weights to be learned.

\subsubsection{Dense Layers}
Dense layers after the concatenation layer add predictive value as they are able to learn interactions between the text features and the clinical predictors. In our multimodal model, the concatenation layer is followed by dense layers with the same sizes of neurons. The output of the concatenation layer is fed into the first dense layer with the $ReLU$ activation function. The output of the second dense layer is then fed into the third dense layer with one neuron. The activation function in this layer is $sigmoid$ with the $binary$ $crossentropy$ loss function.

\section{Experiments}

\subsection{Evaluation Metrics}
To evaluate the classification performance, we use five available metrics: AUC (area under the curve), misclassification rate, precision, recall and F1 score. AUC is the area under the receiver operating characteristic curve which is created by plotting the true positive rate against the false positive rate. Misclassification rate is the proportion of incorrectly classified instances made by a model. Precision is the fraction of relevant instances among the retrieved instances, while recall is the fraction of relevant instances that have been retrieved over the total amount of relevant instances. The F1 score can be interpreted as a weighted average of the precision and recall. The relative contributions of precision and recall to the F1 score are equal. The formula of precision, recall and F1
score can be defined as Eq. 5 ,6, 7:

\begin{equation}\label{eq5}
\text{Precision} = \dfrac{\text{True Positive}}{\text{True Positive + False Positive}}
\end{equation}

\begin{equation}\label{eq6}
\text{Recall} = \dfrac{\text{True Positive}}{\text{True Positive + False Negative}}
\end{equation}

\begin{equation}\label{eq7}
\text{F1 score} = \dfrac{\text{2 * Precision * Recall}}{\text{Precision + Recall}}
\end{equation}

\begin{table}
  \caption{\textbf{Hyperparameter Values}}
  \label{tab:commands}
  \begin{tabular}{lr}
    \toprule
    Parameter & Value\\
    \midrule
    Embedding dimension (d) & 500 \\
    \#neurons in LSTM layer & 100 \\
    CNN filter size & 5 \\
    \#filters in CNN & 128\\
    \#neurons in dense layers & 64 \\
    Dropout rate & 0.2\\
    Recurrent dropout rate & 0.2\\
    Batch size & 64 \\
    \#epochs & 20 \\
    Optimization method & ADAM \\
    
    \bottomrule
  \end{tabular}
\end{table}

\subsection{Experiment Results}
We compare the results of our multimodal BiLSTM model (MI-BiLSTM) to models in four experimental scenarios:
\begin{itemize}
    \item V-NN: A neural network with three dense layers trained on variables from clinical predictors
    \item T-BiLSTM: A BiLSTM model trained on X-ray radiology reports
    \item MI-CNN: The multimodal model using one CNN layer
    \item MI-LSTM: The multimodal model using one LSTM layer
\end{itemize}

Our system is implemented on Keras with a TensorFlow backend \footnote{https://keras.io}. We performed 5-fold cross validation for all experimental scenarios. The hyperparameter settings for training the models are shown in Table 2. These hyperparameters are tuned based on the validation set.

Table 3 presents AUC and misclassification rate of various models on the radiology data set of the UMC Utrecht SMART study. V-NN, which is a neural network with two hidden layers, is trained only on clinical variables. This scenario included age, sex, smoking, systolic blood pressure, diabetes, high-density-lipoprotein cholesterol, total cholesterol, renal function according to the MDRD formula, history of cardiovascular disease stratified for stroke, peripheral artery disease, abdominal aortic aneurysm, and coronary heart disease as predictors and MACE during followup as the outcome in prediction models. T-BiLSTM is trained only on X-ray reports; comparing to V-NN has a lower AUC and a higher misclassification rate. MI-CNN used the framework of the multimodal RNN with a one dimensional convolution layer followed by a max pooling layer, instead of a BiLSTM layer. Table 3 shows the effectiveness of combining both structured and unstructured data in this scenario. MI-LSTM is the multimodal RNN model with one LSTM layer. By comparing MI-LSTM and MI-BiLSTM, it can be seen that using a BiLSTM layer gains the highest metric values. This is because of the ability of the BiLSTM layer as it leverages the knowledge present on both sides of the current word to encode the text.

Figure 2 \footnote{Firatheme version 0.2.1 \cite{vankesteren2020} is used in this plot.} shows the performance of the models using precision, recall and F1 score evaluation metrics. These results are the evidence of the performance of text mining techniques to extract the knowledge in radiology reports and combine them with classical clinical predictors. It is worth mentioning that the recall value of using only clinical variables in V-NN (79.4\%) is higher than the corresponding values for T-BiLSTM (72\%), MI-CNN (73.5\%), and the MI-LSTM model (78.8\%). The F1 score of the V-NN model is also admissible with the value of 77.2\%. This shows the efficiency and relatedness of the laboratory results in predicting the cardiovascular risk. 

\begin{figure*}
  \centering
  \includegraphics[width=\textwidth]{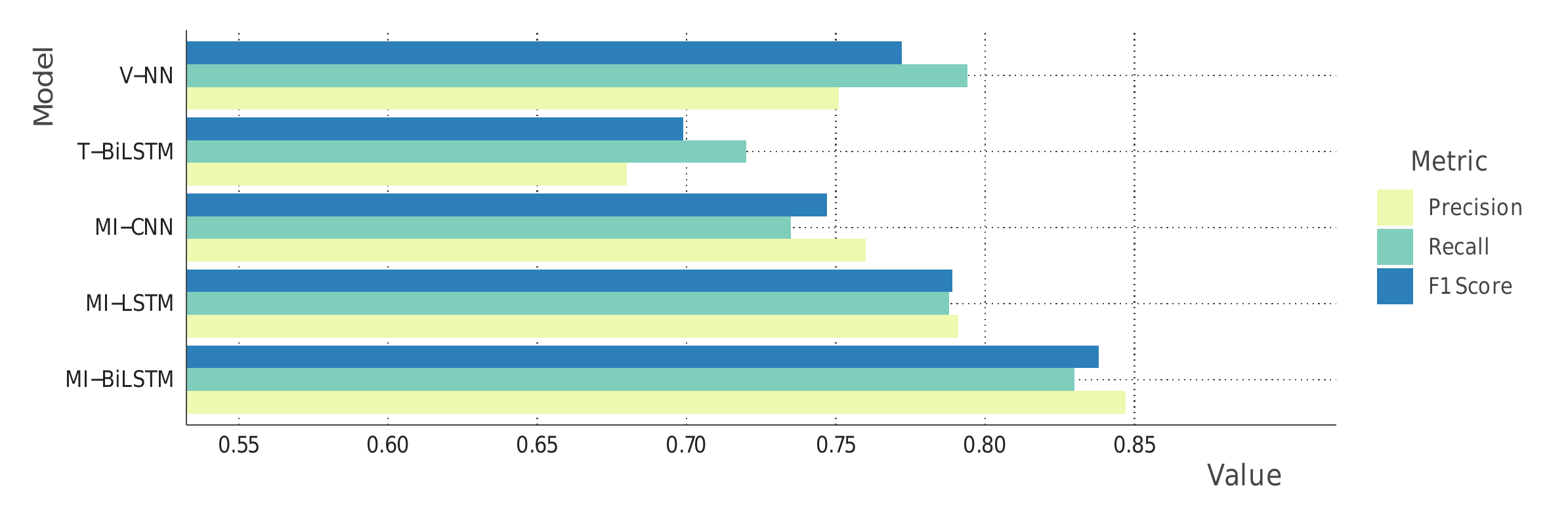}
  \caption[]{\textbf{Performance Comparison of Precision, Recall and F1 Score of the Models}}
\end{figure*}

As can be seen in Figure 2, the RNN models -- MI-LSTM, MI-BiLSTM -- that use both text reports and clinical variables are gained better F1 score values comparing to the CNN-based model and also to the scenarios in which either text reports or clinical variable has been employed. The MI-LSTM and MI-BiLSTM models have gained F1 score of 78.9\% and 83.8\%, respectively. Although CNNs have proven effective for deriving features from sequential data \cite{du2019}, RNNs are specialized for such data and can scale to much longer sequences than would be practical for neural networks without sequence--based specialization \cite{goodfellow2016}.

\begin{table}
  \caption{\textbf{AUC and Misclassification Rate of the Models on Data from the UMC Utrecht SMART Study}}
  \label{tab:commands}
  \begin{tabular}{lcc}
    \toprule
    Model       & AUC   & Misclassification rate\\
    \midrule
    V-NN	    & 0.651	& 0.201 \\
    T-BiLSTM	& 0.570	& 0.300 \\
    MI-CNN	    & 0.730	& 0.214 \\
    MI-LSTM	    & 0.794	& 0.176 \\
    MI-BiLSTM	& 0.847	& 0.143 \\
    \bottomrule
  \end{tabular}
\end{table}

\section{Conclusions}

Text mining methods are the key to successful extraction of clinically important findings from radiology reports. However, previous studies for risk prediction mainly focused on modelling with either structured data or unstructured texts. In this study, we leveraged the knowledge present on both radiology reports and structured laboratory variables to predict cardiovascular risk for patients in UMC Utrecht. We proposed a neural network-driven modelling of radiological language, to integrate clinical and textual features, to supersede traditional algorithms using only clinical variables. To the best of our knowledge, there is no existing work that applies a multimodal RNN-based model to mine free-text radiology reports and combine them with laboratory results for cardiovascular risk prediction. Comparing five different scenarios, the BiLSTM RNN model showed the best performances in cardiovascular risk prediction.

Despite the great potential and the promising results of the deep learning models, the use of these advanced text mining techniques in clinical practice requires support for implementation. Implementation includes the application of the text mining pipeline, and integration in health care process using decision support systems. To help clinicians to interpret the results that come from text mining, collaborations between data scientists, software engineers and clinicians are needed to safeguard technical quality and medical relevance. To this end, in future work we plan to experiment various representations of text data in the text mining framework, and investigate the use of machine learning models with radiology data (reports and demographics) when laboratory results of patients are missing.

The publicly-available source code \footnote{https://github.com/bagheria/CardioRisk-TextMining} of our model can be used to evaluate performance on clinically-relevant classification tasks based on clinical notes and EHR variables.


\begin{acks}
The authors would like to thank Erik-Jan van Kesteren for his comments. 

This work has received support from the EU/EFPIA Innovative Medicines Initiative 2 Joint Undertaking BigData@Heart grant n° 116074. 
The author Folkert Asselbergs is supported by UCL Hospitals NIHR Biomedical Research Centre.

\end{acks}

\bibliographystyle{ACM-Reference-Format}
\bibliography{sample-sigplan}


\end{document}